\documentclass[letterpaper]{article} 
\usepackage{aaai2027}  
\usepackage[hyphens]{url}  
\usepackage{graphicx} 
\usepackage{natbib}  
\usepackage{caption} 
\usepackage{algorithm}
\usepackage{algorithmic}
\usepackage{amsmath}
\usepackage{amssymb}
\usepackage{multirow}
\usepackage{placeins}
\usepackage{tabularx}
\usepackage{array}
\usepackage[table]{xcolor}\definecolor{projectmagenta}{RGB}{150,20,100}

\usepackage{newfloat}
\usepackage{listings}
\DeclareCaptionStyle{ruled}{labelfont=normalfont,labelsep=colon,strut=off} 
\floatstyle{ruled}
\newfloat{listing}{tb}{lst}{}
\floatname{listing}{Listing}

\usepackage{booktabs}

\title{One Skill Does Not Fit All: Automatic Discovery and Taxonomy-Guided Routing of Frame-Selection Skills for Long-Video Question Answering}
\author{
    Jian Hu\textsuperscript{\rm 1}\equalcontrib,
    Zixu Cheng\textsuperscript{\rm 1}\equalcontrib,
    Da Li\textsuperscript{\rm 2},
    Wei Li\textsuperscript{\rm 3},
    Ziquan Liu\textsuperscript{\rm 1},
    Shaogang Gong\textsuperscript{\rm 1}
}
\affiliations{
    \textsuperscript{\rm 1}Queen Mary University of London
    \textsuperscript{\rm 2}Samsung AI Research Institute
    \textsuperscript{\rm 3}Nanyang Technological University, Singapore\\
    jianhu329@gmail.com, zixu.cheng@qmul.ac.uk, dali.academic@gmail.com,\\
    wei.l@ntu.edu.sg, ziquan.liu@qmul.ac.uk, s.gong@qmul.ac.uk\\
    {
        \textcolor{projectmagenta}{
            \url{https://lwpyh.github.io/autoskill_pipeline/}
        }
    }
}

\begin{document}

\nocopyright
\maketitle

\begin{abstract}
Long-Video Question Answering (LVQA) requires locating decisive evidence
in hour-scale videos under a limited frame budget.
Most training-free methods apply the same frame-selection strategy to all
questions, despite substantial variation in the evidence required by
different question types.
Our analysis shows that the relative effectiveness of frame-selection
strategies varies across semantic categories and benchmarks, motivating
adaptive evidence acquisition.
In this paper, we introduce \textbf{AutoSkill}, a source-supervised framework for
automatically discovering and routing executable frame-selection skills.
Starting from a small labelled source pool, LLM agents iteratively propose,
implement, evaluate, and refine candidate skills.
For a target benchmark, AutoSkill uses only unlabelled question and option
text to induce a shared semantic taxonomy, rewrite labelled source examples
into the target style, and estimate a category-to-skill mapping.
Neither target videos nor target answers are used in this process.
At inference time, each question is assigned one skill, which selects the
frames used in a single inference of the frozen video MLLM.
Across five long-video benchmark splits, AutoSkill improves
Qwen2.5-VL-7B and Qwen3.5-4B by 2.4\% and 1.2\%,
respectively, demonstrating the effectiveness of our AutoSkill.
\end{abstract}

\section{Introduction}
Long-Video Question Answering (LVQA) requires answering
questions by locating relevant visual evidence from videos spanning tens of minutes or even hours. Such videos arise in many real-world scenarios,
including lectures, movies, and egocentric recordings
~\cite{zhou2025mlvu,wu2024longvideobench,fu2025video,wang2025lvbench}.
The evidence required to answer a question may appear briefly, recur across distant segments, or depend on the overall evolution of an event.
Exhaustively processing all frames is impractical, as the resulting visual tokens quickly exceed the context capacity of current Multimodal Large
Language Models (MLLMs)~\cite{zhang2024long,chen2025longvila,shu2025video}.
Therefore, effective LVQA requires selecting a small set of informative
frames under a limited visual token budget.

Frame selection is often treated as simple preprocessing. However, the
selected frames define what the frozen MLLM can observe before it answers a
question. Missing evidence cannot be inspected later, while irrelevant
frames use the same limited visual budget. Frame selection therefore acts
as an evidence-acquisition policy in the reasoning pipeline. Different
policies decide how to divide the frame
budget.

Existing approaches address this challenge in two ways. They improve
MLLMs through additional training or reduce visual inputs before inference.
Training-based methods~\cite{zhang2024long,chen2025longvila,shen2024longvu,shu2025video}
extend context windows, compress visual tokens, or learn temporal
representations. Although effective, they require substantial training
resources and are often tied to specific model architectures.
Training-free approaches~\cite{tang2025adaptive,wang2025videotree,sun2025mdp3,
zhang2025q,guo2026logic,zou2025air} provide a more flexible alternative by
selecting informative frames through mechanisms such as query-guided
retrieval, temporal localization, coarse-to-fine search, and iterative
evidence refinement. These methods can be directly applied to frozen MLLMs
with low overhead. However, these approaches rely on a single
manually designed frame-selection strategy, implicitly assuming that one
selection rule can generalise across diverse LVQA questions and target
domains.

\begin{figure*}[ht]
\vspace{-5pt}
  \centering
  \begin{tabular}{@{\hspace{-10pt}} c @{\hspace{0pt}}  c }
    \includegraphics[width=1\columnwidth]{
   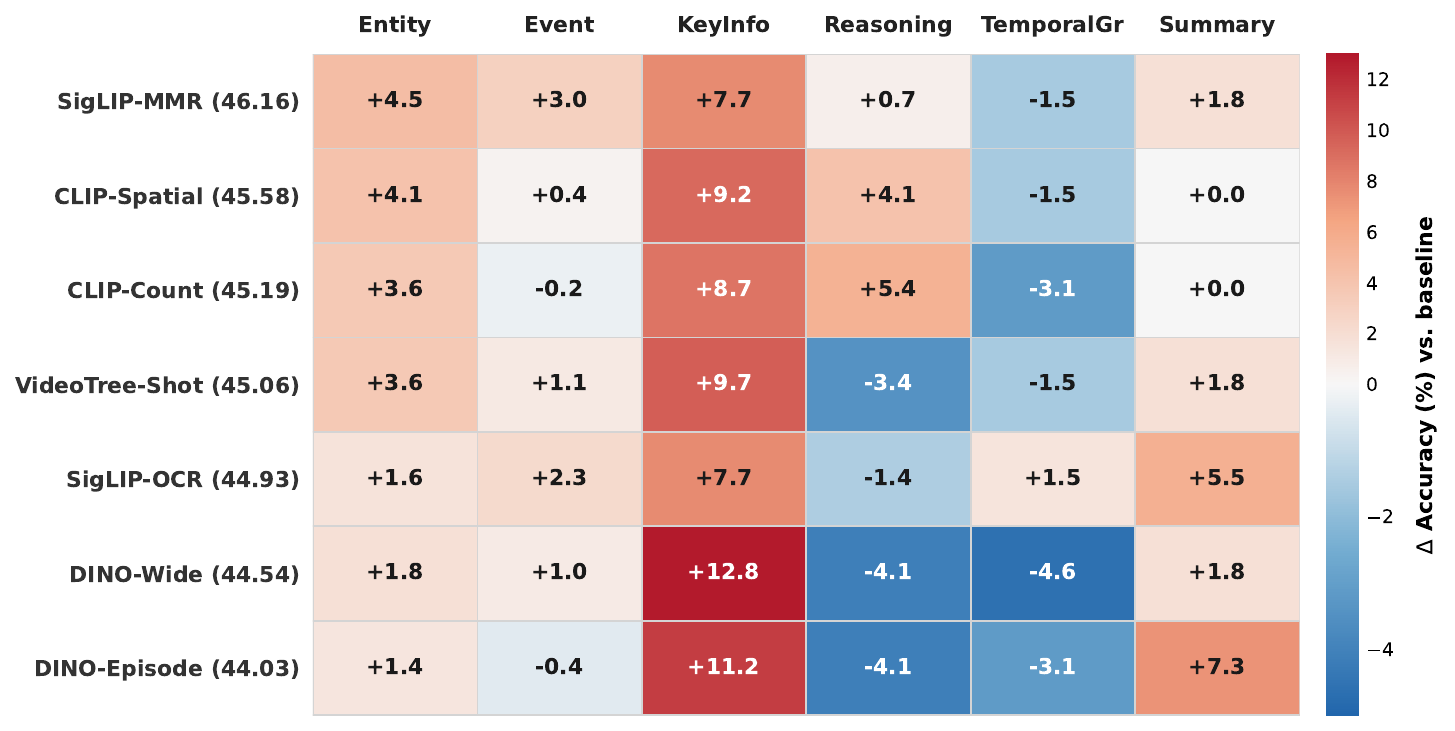}&
      {\includegraphics[width=1\columnwidth]{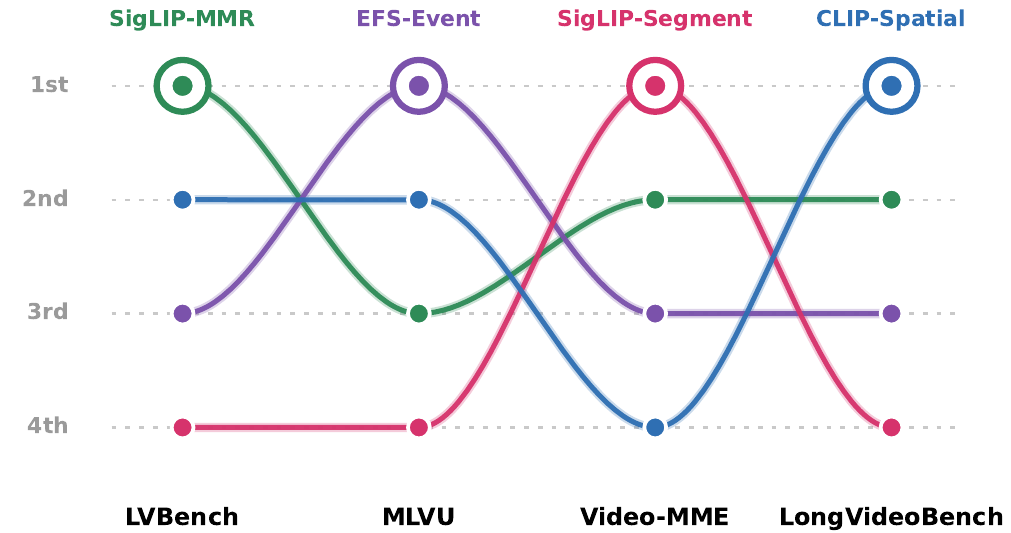}} \\[-0.8ex]
    {\fontsize{9pt}{10pt}\selectfont(a) Per question type accuracy gains ($\Delta$) over uniform sampling on LVBench.} &
     {  \fontsize{9pt}{10pt}\selectfont(b) Relative ranking of the four per-benchmark best sampling skills.}
  \end{tabular}
  \vspace{-2pt}
  \caption{\textbf{Frame-selection policies show two levels of
  heterogeneity.} (a) Within LVBench, different policies are strong on
  different question types. (b) Across benchmarks, their relative rankings
  change. No evaluated fixed policy dominates every category and
  benchmark. These results motivate adaptive evidence acquisition.
  \vspace{-0.3cm}
  \label{fig:moti}}
\end{figure*}

This policy view leads to a deeper finding. LVQA is not one homogeneous
retrieval problem. It contains different evidence-acquisition problems.
For example, temporal reasoning questions may benefit from localized
search, whereas counting or entity-related questions may require broader
coverage or specialized sampling patterns.
Fig.~\ref{fig:moti} shows this heterogeneity at two levels.
Within a benchmark, different question categories favour different
skills.
Across benchmarks, skill rankings also change. This means that evidence
acquisition should adapt to the question and target distribution instead of
using one fixed policy for every case.

This finding raises a practical deployment challenge.
The strongest fixed skill can only be identified by evaluating all
candidates on labelled target examples, which are unavailable beforehand.
A deployable system must therefore infer which skill to use from the
observable target-query distribution alone.
It must also avoid executing every skill for every question, as this would
multiply video inference cost.
The goal is thus to discover effective frame-selection skills, determine
when each should be applied, and execute only one skill per question.

\textbf{AutoSkill} turns this idea into a practical method. AutoSkill represents each evidence-acquisition policy as an executable frame-selection skill.
It automatically discovers complementary frame-selection skills and
performs target-adaptive, taxonomy-guided skill routing.
AutoSkill consists of three stages, as illustrated in
Fig.~\ref{fig:pipeline}.
In Stage~1, AutoSkill uses an LLM agent to automatically discover a
compact toolbox of frame-selection skills.
Starting from a shared labelled source pool, the agent iteratively
proposes, implements, executes, and evaluates candidate selection
programs.
Execution feedback guides subsequent iterations toward more effective
and complementary skills, without requiring manual construction of the
complete skill library.
In Stage~2, AutoSkill adapts the discovered toolbox to a target LVQA
distribution using only unlabelled target question and option text.
It first induces a semantic taxonomy that captures the different forms
of visual evidence required by target queries.
Based on this taxonomy, AutoSkill constructs a target-style routing set
by selecting semantically compatible labelled source examples and
rewriting their questions and options to match the target query style,
while preserving the original video grounding and answer labels.
No target video or target answer annotation enters this adaptation
process.
Evaluating the discovered skills on the adapted routing set estimates
their category-specific utility and produces a category-to-skill
mapping.
In Stage~3, AutoSkill applies this mapping during inference.
Each test question is first assigned to a semantic category, after which
only the corresponding frame-selection skill is executed before the
frozen MLLM generates the final answer.
Unlike post-hoc skill selection, AutoSkill does not require labelled
target evaluation to identify a suitable skill.
Unlike exhaustive skill execution, it retains approximately the
inference cost of a single-skill method.
AutoSkill does use labelled source examples as black-box feedback during skill search and routing-table construction. It is therefore source-supervised, training-free, and target-label-free. It does
not update the video MLLM or auxiliary models used for frame-selection skills. Target
adaptation uses question and option text only. It does not use paired target
videos or target annotations.
\textbf{Our contributions are as follows}:

\begin{figure*}[t]
  \centering
  \includegraphics[width=\textwidth]{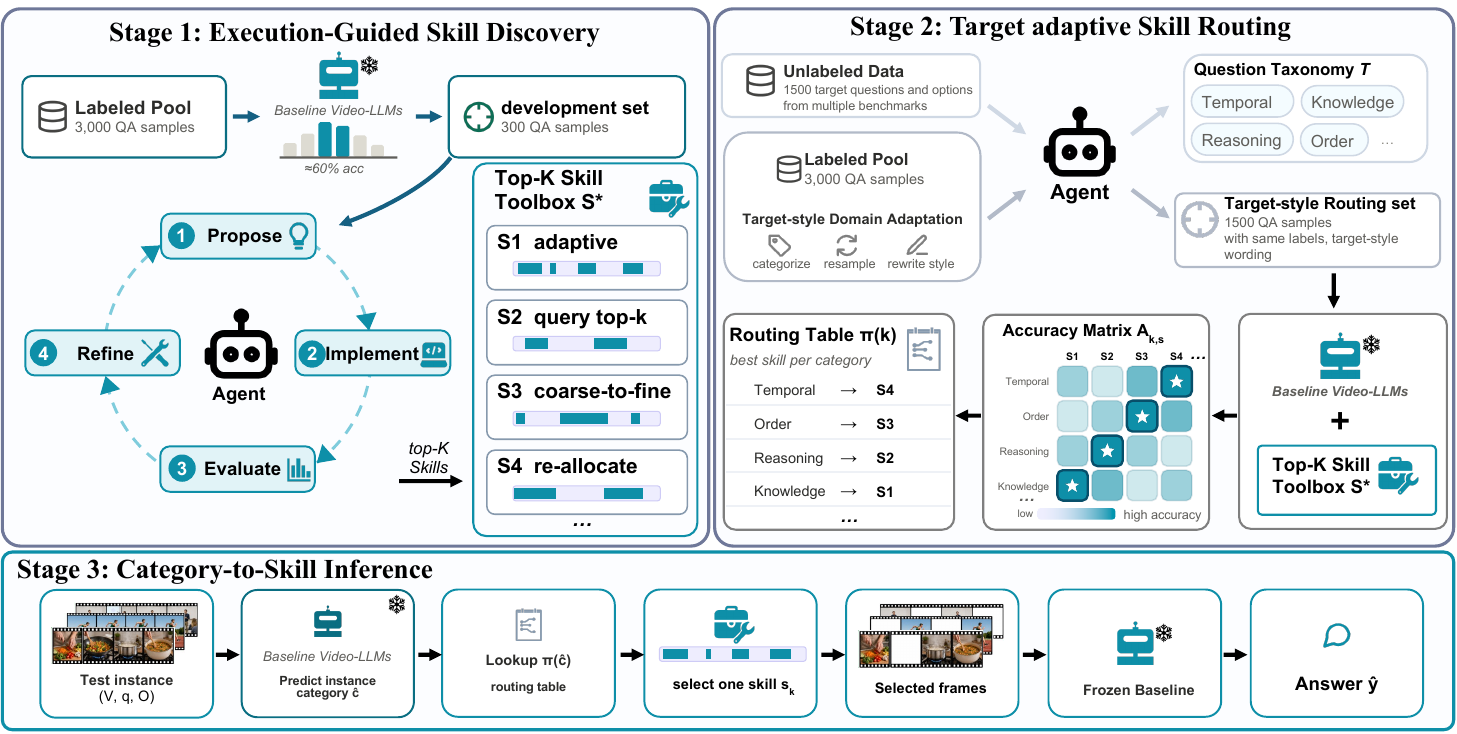}
    \caption{Overview of AutoSkill.
  Stage~1 iteratively discovers a Top-$K$ skill toolbox on a
  capability-focused development set.
  Stage~2 uses unlabelled target question and option text and a target-style
  adaptation of the shared labelled pool to construct a
  category-to-skill routing table.
  Stage~3 predicts the semantic category of each test question, retrieves
  one associated skill, and answers with a single frozen MLLM pass,
  without using target answer annotations.}
  \vspace{-0.3cm}
  \label{fig:pipeline}
\end{figure*}

1. We show that frame selection determines the evidence available to the downstream model. Different frame-selection skills excel on different question categories and benchmarks, indicating that no single policy is universally optimal. This motivates adaptive evidence acquisition.

2. We present \textbf{AutoSkill} as a practical method based on this
finding. It automatically discovers a diverse toolbox of executable frame-selection
skills through LLM-agent-driven exploration and execution feedback.
It further estimates category-specific skill utility from labelled
source examples adapted to the unlabelled target-query distribution,
without accessing target videos or answer annotations during routing-table construction.

3. We introduce an efficient taxonomy-guided router that selects one
automatically discovered frame-selection skill for each test question,
avoiding exhaustive skill execution at inference.
Across four long-video QA benchmarks comprising five evaluation splits,
AutoSkill improves Qwen2.5-VL-7B and Qwen3.5-4B by up to 2.4\% and 1.2\%, respectively.
Without using target answers, it matches or exceeds the post-hoc best
fixed skill selected using target annotations.

\section{Related Work}
\noindent \textbf{Efficient Long-Video Understanding.}
Long-video understanding is constrained by temporal redundancy and limited visual tokens. Training-based methods address this through long-context adaptation, token compression, hierarchical representations, and external memory~\cite{zhang2024long,chen2025longvila,shen2024longvu,shu2025video}, but they usually need video-language supervision and stay tied to one backbone. Training-free methods instead reduce the visual input before inference via query relevance~\cite{zhu2025focus,park2026too,zhang2025q}, visual diversity~\cite{tang2025adaptive,peng2026qca}, event localization~\cite{ye2025re,zou2025air,guo2026logic}, or iterative search~\cite{ye2025re,guo2026logic,park2026too}. These approaches are strong and inexpensive, yet most optimize one selection mechanism and apply it to all questions. Query conditioning can change frame scores, but the evidence-acquisition strategy itself remains fixed. AutoSkill automatically discovers complementary frame-selection skills and learns when each skill should be applied.

\noindent \textbf{Automatic Skill Discovery and Adaptive Routing.}
Adaptive computation appears in mixture of experts~\cite{shazeer2017outrageously,fedus2022switch}, model cascades~\cite{chen2023frugalgpt,ong2024routellm}, and LLM tool use~\cite{schick2023toolformer,qin2024toolllm,patil2024gorilla}. Most systems assume an existing expert or tool library, and their routers often need supervised trajectories, learned retrievers, reward optimization, or multi-expert execution~\cite{ong2024routellm,zheng2026skillrouter,zou2025autotool,chen2025symbolic}. Agent systems can request and refine tools during solving~\cite{fei2025mcp,zou2025autotool}, but rarely build a reusable video skill toolbox from downstream evaluation. AutoSkill brings automatic skill discovery into long-video frame selection. An LLM agent discovers frame-selection skills on a capability-focused development set. AutoSkill then distills their execution history into a category-to-skill mapping without parameter updates or target benchmark answers.

\section{Methodology}

\subsection{Frame Selection as Evidence Acquisition}

Long-Video Question Answering (LVQA) requires selecting informative visual
evidence from long videos before feeding them into a frozen MLLM.
Given a question $q$, candidate options $\mathcal O$, and a video $V$,
a frame-selection skill $s$ selects a subset of frames:
\begin{equation}
\mathcal F_s=s(V,q),
\qquad
|\mathcal F_s|\leq B,
\end{equation}
where $B$ denotes the frame budget.
The frozen MLLM then produces the answer:
\begin{equation}
\hat y_s=f_{\theta}(\mathcal F_s,q,\mathcal O).
\end{equation}
We use \emph{policy} to mean a general strategy for acquiring visual
evidence. We use \emph{skill} to mean an executable program that implements
such a policy. Thus, each frame-selection skill $s$ implements one
evidence-acquisition policy and controls which visual evidence reaches the
frozen MLLM.
Existing training-free LVQA methods typically assume a universal frame
selection strategy.
However, different questions may require different evidence acquisition
patterns, motivating a toolbox of complementary skills:
$
\mathcal S=\{s_1,\ldots,s_L\}.
$
For a target distribution $\mathcal D_t$, the utility of a fixed skill is
defined as:
\begin{equation}
U_t(s)
=
\mathbb E_{(V,q,\mathcal O,y)\sim\mathcal D_t}
\left[
\mathbb I
(
f_{\theta}(s(V,q),q,\mathcal O)=y
)
\right].
\end{equation}
Let $U_t(s\mid c)$ denote the utility of skill $s$ for semantic category
$c$. Evidence acquisition is heterogeneous when the skill with the highest
utility changes across categories or target distributions. A universal
selector uses one skill for every category. An adaptive selector instead
uses a mapping from categories to skills.
The strongest fixed skill is therefore:
\begin{equation}
s_t^{*}
=
\arg\max_{s\in\mathcal S}U_t(s),
\end{equation}
However, $s_t^{*}$ cannot be identified before deployment because its
utility requires target videos and answer annotations for evaluation.
Moreover, the optimal skill may vary across target distributions.
Therefore, a practical LVQA system should select skills according to the
target query distribution without exhaustively evaluating all skills.
AutoSkill addresses this problem under a transductive target-adaptation
setting.
It is given a labelled source pool
$
\mathcal D_s
=
\{(V_i^s,q_i^s,\mathcal O_i^s,y_i^s)\}_{i=1}^{N_s}
$
and unlabelled target question and option pairs:
$
\mathcal Q_t
=
\{(q_i^t,\mathcal O_i^t)\}_{i=1}^{N_t}.
$
Only target query semantics are available during adaptation, while target
videos, target answers, and target-domain skill performance are not used.

AutoSkill first discovers a toolbox $\mathcal S^{*}$ from source data and
then constructs a category-to-skill mapping
\begin{equation}
\pi(k)\in\mathcal S^{*},
\qquad k\in\{1,\ldots,K_q\},
\end{equation}
where $K_q$ denotes the number of semantic query categories.
At inference time, only the skill assigned to the predicted category is
executed.

This setting is source-supervised because source answer labels are used to
score candidate programs and construct the routing table. It is
training-free because these evaluations do not update any MLLM model or auxiliary models used in skills. It is also target-label-free because target
annotations are not used. During target adaptation, AutoSkill uses target
question and option text but not paired target videos or ground-truth answers.

\subsection{Capability-Focused Development Set Construction}
\label{sec:cap-focused}

Automatic discovery of frame-selection skills requires an informative
evaluation environment.
Randomly sampled LVQA examples often provide limited feedback because they
may be either trivially solved or consistently failed by the frozen MLLM,
making it difficult to distinguish the benefits of different selection
strategies.

We therefore construct a capability-focused development set from a shared
labelled source pool.
Specifically, we collect $N_s=3,000$ examples from
VideoVista~\cite{li2024videovista},
ALLVB~\cite{tan2025allvb}, and
LongVideo-Reason~\cite{chen2026scaling}.
The source pool is disjoint from all target evaluation benchmarks and is
only used for skill discovery and routing construction.
From this pool, we select intermediate-difficulty examples:
\begin{equation}
\mathcal D_{\mathrm{dev}}
=
\{(V_i^s,q_i^s,\mathcal O_i^s,y_i^s)\}_{i=1}^{N_{\mathrm{dev}}},
\end{equation}
where $N_{\mathrm{dev}}=300$ and the frozen MLLM achieves approximately
60\% accuracy.
Such examples provide both successful and failed cases, enabling the agent
to discover when alternative frame-selection mechanisms are beneficial.
To maintain diversity, we uniformly stratify examples according to video
duration and question categories.
We use $300$ examples to balance discovery quality and repeated evaluation
cost, with the impact of development-set size analysed in
Tab.~\ref{tab:additional_comparisons}(c).

\subsection{Execution-Guided Skill Discovery}
\label{sec:agent-workflow}

Without prior knowledge of effective frame-selection strategies, AutoSkill uses an LLM agent to discover executable skills rather than manually designing a fixed toolbox.

Starting from $\mathcal D_{\mathrm{dev}}$, the agent iteratively analyses
previous execution results and failure cases, proposes new selection
mechanisms, implements executable programs, and evaluates them using the
frozen LVQA pipeline.
The execution feedback guides subsequent exploration toward more effective
and complementary evidence acquisition strategies.
Each candidate skill specifies its selection hypothesis, input signals,
auxiliary models if required, and frame allocation procedure.
Before evaluation, a reviewer agent verifies program correctness, auxiliary
model constraints, and compliance with the frame budget.
All candidates are evaluated under the same frozen MLLM and inference
configuration, ensuring that performance differences mainly reflect the
frame-selection strategy.
For each skill $s$, we measure:
\begin{equation}
R(s,\mathcal D_{\mathrm{dev}})
=
\frac{1}{N_{\mathrm{dev}}}
\sum_i
\mathbb I
[
f_{\theta}(s(V_i^s,q_i^s),q_i^s,\mathcal O_i^s)=y_i^s
].
\end{equation}
Besides overall accuracy, the agent receives stratified feedback by video duration, question category, and data source to identify where each skill is effective. After multiple exploration cycles, the Top-$K_s$ remaining skills form the discovered toolbox, here we set $K_s=5$:
\begin{equation}
\mathcal S^{*}=\{s_1,\ldots,s_{K_s}\}.
\end{equation}
The toolbox is then frozen before target-query adaptation.

\subsection{Target-Adaptive Skill Routing}
\label{sec:skill-routing}

Although the discovered toolbox contains complementary skills, their utility on the target distribution is unknown. AutoSkill therefore estimates category-specific skill preference from target query semantics and source supervision.

\noindent\textbf{Query Taxonomy Induction.}
We sample $N_q$ unlabelled question--option pairs from each of the $J=5$ target benchmark splits and merge them into a unified query set:
\begin{equation}
\mathcal Q_t
=
\bigcup_{j=1}^{J}
\left\{
(q_{j,i}^{t},\mathcal O_{j,i}^{t})
\right\}_{i=1}^{N_q}.
\end{equation}
We set $N_q=300$, with the effect of query-set size analysed in Tab.~\ref{tab:routing_size}. An LLM agent jointly analyses the pooled queries and organizes them into a shared semantic taxonomy:
\begin{equation}
\mathcal T
=
\{\tau_1,\ldots,\tau_{K_q}\},
\qquad K_q=19.
\end{equation}
The induced categories characterize the evidence required to answer each query, including temporal localization, event ordering, counting, and global understanding.
\textbf{Full taxonomy details are provided in the supplementary material.}

\noindent\textbf{Target-Style Routing Set Construction.}
The taxonomy reveals what evidence is required by target queries,
but it does not reveal which discovered skill is suitable for each category.
Therefore, we construct a source-grounded routing set.
For each category $\tau_{k}$, we retrieve semantically compatible source
examples and rewrite their questions and options with an LLM to match the target query
style, while preserving the original video, evidence requirement, and answer
label:
\begin{equation}
\mathcal D_{\mathrm{route}}
=
\{
(V_n^s,\widetilde q_n,
\widetilde{\mathcal O}_n,y_n^s,c_n)
\}_{n=1}^{N_{\mathrm{route}}},
\end{equation}
The routing set is constructed from source examples disjoint from
$\mathcal D_{\mathrm{dev}}$:
\begin{equation}
\mathcal D_{\mathrm{dev}}\cap
\mathcal D_{\mathrm{route}}
=
\varnothing,
\end{equation}
Therefore, $\mathcal D_{\mathrm{route}}$ provides a target-style proxy for
comparing skills without accessing target videos or answers.

\begin{table*}[t]
\caption{Comparison on long video understanding benchmarks. $^{\dagger}$ marks Qwen2.5-VL-7B post-training methods whose scores are taken from ParaVT~\cite{yang2026paravt}. $^{\ddagger}$ denotes our reimplementation under the 128-frame budget. Each skill is a fixed training-free selector, and AutoSkill routes among them.}
\centering
\vspace{-10pt}
\setlength{\tabcolsep}{4pt}
\begin{tabular}{l c cc c c c c}
\toprule
\multirow{2}{*}{Model} & \multirow{2}{*}{Size}
& \multicolumn{2}{c}{MLVU}
& LongVideoBench
& \multirow{2}{*}{VideoMME}
& \multirow{2}{*}{LVBench}
& \multirow{2}{*}{Avg/$\Delta$} \\
\cmidrule(lr){3-4}
&
& dev & test
& val
& & & \\
\midrule

\multicolumn{8}{c}{\textit{Proprietary Models}} \\\hline
GPT-4o~\cite{hurst2024gpt} & -
& 64.6 & 54.9 & 66.7 & 71.9 & 48.9 & -- \\

Gemini-2.5-Pro~\cite{comanici2025gemini}
& - & 81.2 &  -
& -
& 84.3
& 78.7
& -- \\

\midrule

\multicolumn{8}{c}{\textit{Open-Source VLMs}} \\\hline
Video-LLaVA~\cite{lin2024video} & 7B
& 36.2 & 30.7
& 37.6
& 39.9
& -
& -- \\

LLaVA-OneVision~\cite{li2024llava} & 7B
& 64.7 & 47.2
& 56.4
& 58.3
& -
& -- \\


Video-XL~\cite{shu2025video} & 7B
& 64.9 & 45.5
& 50.7
& 55.5
& -
& -- \\


Kangaroo~\cite{liu2026kangaroo} & 8B
& 61.0 & -
& 54.8
& 56.0
& 39.4
& -- \\

LongVU~\cite{shen2024longvu} & 7B
& 65.4 & -
& -
& 60.6
& -
& -- \\

LongVA~\cite{zhang2024long} & 7B
& 56.3 & 41.1
& -
& 52.6
& -
& -- \\

LongVILA~\cite{chen2025longvila} & 7B
& - & -
& 57.1
& 60.1
& -
& -- \\

Qwen2.5-VL~\cite{bai2025qwen2} & 7B
&  64.4 & 50.7
& 58.7
& 64.0
& 42.3
& -- \\

LongVILA-R1~\cite{chen2026scaling} & 7B
& - & -
& 57.6
& 62.4
& -
& -- \\

Video-R1~\cite{feng2026video} & 7B
& 65.0 & 49.2
& 52.0
& 61.1
& 38.7
& -- \\

VideoZoomer~\cite{ding2025videozoomer} & 7B & 68.8  & \textbf{55.8} & 57.7 & 65.2 & 41.5 & -- \\

VideoChat-R1$^{\dagger}$~\cite{li2025videochat} & 7B
& 58.7 & -
& 49.2
& 50.4
& 23.8
& -- \\

ReWatch-R1$^{\dagger}$~\cite{zhang2025rewatch} & 7B
& 60.1 & -
& 53.6
& 58.8
& 38.5
& -- \\
Video-Thinker$^{\dagger}$~\cite{wang2025video} & 7B
& 65.2 & -
& 56.0
& 61.9
& -
& -- \\
Conan$^{\dagger}$~\cite{ouyang2026conan} & 7B
& 59.2 & -
& 54.5
& 55.5
& 38.2
& -- \\
LongVT-RFT$^{\dagger}$~\cite{yang2026longvt} & 7B
& 59.4 & -
& 54.7
& 59.5
& 37.9
& -- \\
SAGE$^{\dagger}$~\cite{jain2026sage} & 7B
& 49.7 & -
& 37.4
& 44.1
& 31.8
& -- \\

\midrule
\multicolumn{8}{c}{\textit{Discovered Skill Toolbox (Ours, only selected Top-5 skills)$^{\ddagger}$}} \\\hline
Qwen2.5-VL$^{\ddagger}$~\cite{bai2025qwen2} & 7B & 66.0 & 47.7 & 61.2 & 64.6 & 42.5 & 56.4/-- \\
SigLIP-MMR & 7B & 68.3 & 51.5 & 61.6 & \textbf{65.3} & 46.2 & 58.6/+2.2 \\
CLIP-Spatial & 7B & 68.8 & 52.1 & \textbf{62.0} & 64.0 & 45.6 & 58.5/+2.1 \\
CLIP-Count & 7B & 67.7 & 50.6 & \textbf{62.0} & 64.1 & 45.2 & 57.9/+1.5 \\
CLIP-MMR & 7B & 67.6 & 51.2 & 61.1 & 64.5 & 44.9 & 57.9/+1.5 \\
SigLIP-OCR & 7B & 67.4 & 50.1 & 61.5 & \textbf{65.3} & 44.9 & 57.8/+1.4 \\
\hline
AutoSkill (ours)$^{\ddagger}$ & 7B
& \textbf{69.0} & 51.1 & 61.9 & 65.2 & \textbf{46.8}
& \textbf{58.8/+2.4} \\
\hline
\end{tabular}
\vspace{-5pt}
\label{tab:long_video}
\end{table*}
\noindent\textbf{Category-to-Skill Assignment.}
For each category $k$ and skill $s$, we compute:
\begin{equation}
A_{k,s}^{\mathrm{route}}
=
\frac{1}{N_k}
\sum_{n\mid c_n=k}
\mathbb I
[
f_{\theta}
(
s(V_n^s,\widetilde q_n),
\widetilde q_n,
\widetilde{\mathcal O}_n
)
=
y_n^s
],
\end{equation}
The router assigns each category to the skill with the highest estimated
utility:
\begin{equation}
\pi(k)
=
\arg\max_{s\in\mathcal S^{*}}
A_{k,s}^{\mathrm{route}}.
\end{equation}
Importantly, this utility estimation relies only on source videos and
labels.
It serves as a proxy for category-specific skill preference rather than
measured target performance.
Several categories can share one skill. In the Qwen2.5-VL mapping,
counting and anomaly detection select SigLIP-MMR, temporal ordering and
causal reasoning select CLIP-Spatial, and timestamp-specific questions
select SigLIP-OCR. \textbf{The full category taxonomy, rewrite prompts, and mapping table are
provided in the supplemental material.}

\subsection{Inference with Category-to-Skill Routing}
At inference time, AutoSkill removes benchmark templates, answer choices,
and output instructions from each query. Let $\bar q$ denote the remaining
question content. AutoSkill assigns $\bar q$ to a semantic category and
retrieves its associated skill:
\begin{equation}
\hat c=A_{\mathrm{cat}}(\bar q,\mathcal T),
\qquad
\hat s=\pi(\hat c).
\end{equation}
The routed skill then selects a fixed-budget frame set
$\hat{\mathcal F}=\hat s(V,q)$, from which the frozen MLLM predicts
\begin{equation}
\hat y=f_{\theta}(\hat{\mathcal F},q,\mathcal O).
\end{equation}
This category-level routing matches frame selection to each query's evidence
while executing only one skill.
\section{Experiments}
\subsection{Experimental Setup}

\noindent \textbf{Benchmarks.} We evaluate on four long-video QA benchmarks
covering five splits. They are MLVU dev and
test~\cite{zhou2025mlvu}, LongVideoBench val~\cite{wu2024longvideobench},
Video-MME without subtitles~\cite{fu2025video}, and
LVBench~\cite{wang2025lvbench}. We use their official evaluation
protocol and a unified frame budget of $B{=}128$.

\begin{table}[!t]
\caption{
Results of the Qwen3.5-4B baseline, Top-5 discovered skills,
and AutoSkill.
$^{\dagger}$ denotes our reimplementation.
LongVB denotes LongVideoBench.
}
\label{tab:qwen35_skills}
\centering
\vspace{-5pt}
\fontsize{7.5pt}{9pt}\selectfont
\renewcommand{\arraystretch}{1.02}
\setlength{\tabcolsep}{1.2pt}

\begin{tabularx}{\columnwidth}{
@{}>{\raggedright\arraybackslash}X
cc
c
c
c
c@{}
}
\toprule
\multirow{2}{*}{\textbf{Method / Skill}}
& \multicolumn{2}{c}{\textbf{MLVU}}
& \textbf{LongVB}
& \textbf{VideoMME}
& \textbf{LVBench}
& \textbf{Avg./$\Delta$} \\
\cmidrule(lr){2-3}
& \textbf{dev}
& \textbf{test}
& \textbf{val}
& 
& 
& \\
\midrule

Qwen3.5$^{\dagger}$
& 69.9
& 55.1
& 61.3
& 67.3
& 42.3
& 59.2/-- \\

\midrule

QMSD-Zoom
& 69.5
& 55.8
& 62.2
& 65.8
& 44.6
& 59.6/+0.4 \\

Boundary-Evidence
& 69.2
& 55.4
& 62.1
& 64.9
& 45.1
& 59.3/+0.1 \\

Indexed-Evidence
& 68.9
& 55.5
& 62.3
& 66.1
& 46.3
& 59.8/+0.6 \\

Clip-Face-Expression
& 69.9
& 55.5
& 62.3
& 66.0
& 46.1
& 60.0/+0.8 \\

Count-Diversity-Nudge
& 69.9
& 55.4
& 62.4
& 67.0
& 42.3
& 59.4/+0.2 \\

\midrule

\textbf{AutoSkill}
& 69.4
& \textbf{57.5}
& \textbf{62.4}
& \textbf{67.1}
& 45.4
& \textbf{60.4/+1.2} \\

\bottomrule
\end{tabularx}
\vspace{-10pt}
\end{table}

\noindent \textbf{Implementation details.} 
Routing uses $J{=}5$ evaluation splits with $M{=}300$ unlabelled question and option pairs from each split. This gives $|\mathcal{Q}_{\mathrm{tgt}}|{=}1{,}500$ target questions. Only their question and option text is used during routing-table construction. Paired target videos and annotations are not used. The agent induces $K_q{=}19$ categories, and builds a rewritten routing set of $N{=}1{,}500$ samples from the $P{=}3{,}000$ labelled source pool in Section~\ref{sec:cap-focused}. Source answer labels are used to score skills, but no parameters are updated. Backbones are frozen Qwen2.5-VL-7B~\cite{bai2025qwen2} and Qwen3.5-4B~\cite{qwen3.5} and a direct MCQ prompt. Discovery is run by Sonnet 5 with Auto Research~\cite{yang2026aris}, with a GPT-5.5 reviewer who checks proposals before coding. Skills may call auxiliary models of at most 1B parameters, including CLIP, SigLIP, DINOv2, and Grounding-DINO. They may change only frame allocation. Following Table~\ref{tab:abl_components}, we run $T{=}6$ cycles with $M_c{=}5$ evaluations per cycle. Cycle~1 proposes five skills. Later cycles use three refinements plus two explorations. Failed reviews are dropped before GPU runs. Runtime errors fall back to uniform sampling. Skills that regress against uniform are excluded from $\mathcal{S}^{*}$. No human filtering or editing is applied. Detailed prompts and the class-skill mapping are in the supplemental material.


\subsection{Experimental Results}

Figure~\ref{fig:moti} provides direct evidence for heterogeneous acquisition
policies: skills exhibit distinct category-wise strengths on LVBench, and
their rankings vary across benchmarks. This observation motivates AutoSkill.
We next examine whether automatic discovery and routing can translate such
complementarity into improved LVQA performance.

\noindent\textbf{Results with Qwen2.5-VL-7B.}
Table~\ref{tab:long_video} compares AutoSkill with the frozen backbone,
individual discovered skills, and existing proprietary and open-source
models. Under the unified 128-frame budget, the baseline averages 56.4\%
across five benchmark splits. All Top-5 skills improve the average by
1.4--2.2\%, showing that discovery consistently produces effective
strategies. Their strengths are complementary: CLIP-Spatial performs best on
both MLVU splits and ties for the best LongVideoBench result among individual
skills, while SigLIP-MMR performs best on LVBench and ties on VideoMME.
By routing queries among these skills, AutoSkill achieves 69.0\%, 51.1\%,
61.9\%, 65.2\%, and 46.8\% on MLVU-dev, MLVU-test, LongVideoBench,
VideoMME, and LVBench, respectively. It obtains the best overall average
among the Qwen2.5-VL variants, improving the baseline by 2.4\% and every
individual split. These results show that taxonomy-guided routing converts
skill complementarity into consistent overall gains.

\noindent\textbf{Results with Qwen3.5-4B.}
Table~\ref{tab:qwen35_skills} repeats discovery with frozen Qwen3.5-4B,
yielding a separate Top-5 toolbox rather than transferring skills from
Qwen2.5-VL. The baseline averages 59.2\%; individual skills improve it by
0.1--0.8\%, while AutoSkill reaches 60.4\% ($+1.2\%$), showing
that the discovery-and-routing protocol adapts to different models.

\noindent\textbf{Learned category-to-skill mapping.}
The routing table assigns the 19 categories based on
category-specific accuracy on the routing set.
SigLIP-MMR covers categories requiring diverse evidence,
such as counting, event identification, and spatial location, whereas
CLIP-Spatial handles categories involving co-occurring evidence, including
causal reasoning and temporal ordering.
SigLIP-OCR addresses action recognition and timestamp-specific
questions, while narrative-plot questions retain uniform sampling for
temporal coverage.
\subsection{Ablation Studies}

\begin{table}[t]
\centering
\caption{Skill discovery across successive feedback cycles on the development set. $\Delta$ is relative to uniform sampling. We use $T{=}6$ in the main experiments after observing saturation.}
\label{tab:abl_components}
\vspace{-5pt}
\small
\setlength{\tabcolsep}{4pt}
\begin{tabular}{p{4.8cm}cc}
\toprule
Variant & Avg.\ acc.\ (\%) & $\Delta$ \\
\midrule
Baseline                     & 60.0 & -- \\
Best skill after Cycle 1                & 61.0 & +1.0 \\
Best skill after Cycle 4                & 62.7 & +2.7 \\
Best skill after Cycle 6                & 63.0 & +3.0 \\
Best skill after Cycle 8                & 63.0 & +3.0 \\
\bottomrule
\end{tabular}
\vspace{-10pt}
\end{table}

\noindent\textbf{Execution feedback accumulates useful refinements.}
As shown in Table~\ref{tab:abl_components}, the best skill after Cycle~1 improves Qwen2.5-VL-7B by 1.0\%. Continued diagnosis and refinement raise the gain to 2.7\% after Cycle~4 and 3.0\% after Cycle~6. Extending the search to Cycle~8 yields no further improvement on the development set. Execution feedback therefore continues to help through early cycles and then saturates, so we set $T{=}6$ for the main experiments.

\noindent\textbf{Comparison with training-free frame selection.}
Table~\ref{tab:additional_comparisons}(a) compares AutoSkill with
recent training-free frame-selection methods under the same
Qwen2.5-VL-7B backbone and 32-frame budget.
AutoSkill achieves 67.9\% on MLVU-dev and 61.5\% on
LongVideoBench, outperforming mDP$^{3}$~\cite{sun2025mdp3}
and A.I.R.~\cite{zou2025air} on both benchmarks.
Unlike these fixed frame-selection strategies, AutoSkill
automatically discovers a toolbox of complementary
evidence-acquisition policies and routes each query to an
appropriate skill while retaining a single-skill inference budget.
\newlength{\subtableAheight}
\newlength{\subtableBheight}
\newlength{\subtableCheight}

\setlength{\subtableAheight}{3.15cm}
\setlength{\subtableBheight}{3.15cm}
\setlength{\subtableCheight}{3.15cm}

\begin{table*}[t]
\centering
\caption{
(a) Comparison with other training-free frame-selection methods under the 32-frame selection budget.
(b) Computational cost versus RL post-training; data denote
SFT+RL samples for the baselines and labelled source plus
unlabelled target-text samples for AutoSkill.
$^{\dagger}$ uses training time reported by the official repository,
and $^{\ddagger}$ includes discovery and router built.
(c) Development-set analysis across different capability-focused
configurations; Base and Best denote uniform-sampling and
best-discovered-skill accuracy, respectively.
LongVB denotes LongVideoBench.
}
\label{tab:additional_comparisons}
\vspace{-5pt}

\scriptsize

\begin{minipage}[t][\subtableAheight][t]{0.37\textwidth}
\vspace{0pt}
\centering

{\small\bfseries
(a) Frame Selection Method Comparison
\par}
\vspace{3pt}

\renewcommand{\arraystretch}{0.67}
\setlength{\tabcolsep}{1pt}

\begin{tabular*}{\linewidth}{
@{\extracolsep{\fill}}lccc@{}
}
\toprule
\textbf{Method} & \textbf{Frames}
& \textbf{MLVU-dev}
& \textbf{LongVB} \\
\midrule

Qwen2.5-VL & 32 
& 59.3
& 58.1 \\

\quad + mDP$^{3}$~\cite{sun2025mdp3}
& 32 & 66.2
& 60.0\\

\quad + A.I.R.~\cite{zou2025air}
& 32 & 67.5
& 61.4 \\

\midrule

\quad + AutoSkill
& 32 & \textbf{67.9}
& \textbf{61.5} \\

\bottomrule
\end{tabular*}

\end{minipage}
\hfill
\begin{minipage}[t][\subtableBheight][t]{0.43\textwidth}
\vspace{2pt}
\centering

{\small\bfseries
(b) Computational cost versus RL post-training
\par}
\vspace{1pt}

\renewcommand{\arraystretch}{0.77}
\setlength{\tabcolsep}{1.3pt}

\begin{tabular*}{\linewidth}{
@{\extracolsep{\fill}}lccc@{}
}
\toprule
\textbf{Method}
& \textbf{Update}
& \textbf{Data}
& \textbf{Cost} \\
\midrule

Video-R1~\cite{feng2026video}
& Yes
& 165K+260K
& 440 A100-h \\

LongVILA-R1~\cite{chen2026scaling}
& Yes
& 36K+170K
& 9.2--12.3K A100-h$^{\dagger}$ \\

VideoZoomer~\cite{ding2025videozoomer}
& Yes
& 52K
& $\sim$768 H100-h \\

AutoSkill
& No
& 3K+1.5K
& $\sim$40-45 A100-h$^{\ddagger}$ \\

\bottomrule
\end{tabular*}

\end{minipage}
\hfill
\begin{minipage}[t][\subtableCheight][t]{0.17\textwidth}
\vspace{0pt}
\centering

{\small\bfseries
(c) Development Set
\par}
\vspace{3.5pt}

\renewcommand{\arraystretch}{1.0}
\setlength{\tabcolsep}{1pt}

\begin{tabular*}{\linewidth}{
@{\extracolsep{\fill}}cccc@{}
}
\toprule
\textbf{$N_{\mathrm{dev}}$}
& \textbf{Base}
& \textbf{Best}
& \textbf{$\Delta$} \\
\midrule

300
& 60.0
& 63.0
& +3.0 \\

450
& 54.4
& 58.4
& +4.0 \\

1{,}000
& 49.9
& 52.9
& +3.0 \\

\bottomrule
\end{tabular*}

\end{minipage}
\vspace{-30pt}
\end{table*}


\noindent\textbf{Efficiency analysis.}
Table~\ref{tab:additional_comparisons}(b) compares AutoSkill with
RL-based video post-training. AutoSkill requires only about 30
discovery evaluations and router built (roughly 340-45 A100-h), approximately
$11\times$ less training time than the 440 A100-h reported by
Video-R1~\cite{feng2026video}. Unlike RL methods, which rely on
large-scale SFT and RL training data, AutoSkill uses only labelled
source data and unlabelled target queries for skill discovery and
routing. Under the matched protocol, it improves the frozen backbone
by $+2.4\%$ without weight updates, while inference requires only one
video-MLLM pass and a lightweight text-only routing call.

\noindent\textbf{Development-set size.}
Table~\ref{tab:additional_comparisons}(c) compares discovery on capability-focused sets of 300, 450, and 1{,}000 samples. Absolute baseline accuracy decreases as the set grows harder, but the best discovered skill improves the baseline by $+3.0\%$, $+4.0\%$, and $+3.0\%$, respectively. Enlarging the development set therefore does not yield a substantially larger discovery gain, while evaluation cost scales roughly with $N_{\mathrm{dev}}$. We use $N_{\mathrm{dev}}{=}300$ in the main experiments.

\begin{table}[t]
\centering
\caption{
\textbf{Comparison of fixed selection, ensembling, and adaptive
routing.}
The target-adapted global fixed strategy selects one skill
for all queries using the same target-style routing set as
AutoSkill.
The post-hoc best fixed skill is selected using target
annotations and is included only as an oracle reference.
}
\label{tab:router_comparison}
\vspace{-4pt}

\small
\renewcommand{\arraystretch}{1.0}
\setlength{\tabcolsep}{2pt}

\begin{tabularx}{\columnwidth}{
@{}>{\raggedright\arraybackslash}Xcccc@{}
}
\toprule
\textbf{Strategy}
& \shortstack{\textbf{Target} \textbf{labels}}
& \textbf{Avg. acc.}
& \textbf{$\Delta$}
& \textbf{Passes} \\
\midrule
Uniform sampling
& No
& 56.4
& --
& 1 \\

Discovery Cycle Router
& No
& 57.4
& +1.0
& 1 \\

Target-adapted global fixed
& No
& 57.9
& +1.5
& 1 \\

Majority voting
& No
& 58.6
& +2.2
& 5 \\

Post-hoc best fixed skill
& Yes
& 58.6
& +2.2
& 1 \\

Optimal category mapping
& Yes
& 60.4
& +4.0
& 1 \\

AutoSkill router
& No
& \textbf{58.8}
& \textbf{+2.4}
& 1 \\
\bottomrule
\end{tabularx}

\vspace{-6pt}
\end{table}

\noindent\textbf{Router comparison.}
Table~\ref{tab:router_comparison} separates the gains from
target-query adaptation and category-specific routing.
Using the same target-style routing set, the target-adapted
global fixed strategy selects one skill for all target queries
and improves the uniform baseline by 1.5 points.
AutoSkill instead routes each semantic category to its
associated skill, reaching 58.8\% and providing a further
0.9-point gain under the same supervision and inference
budget. This shows that the improvement comes not only from
identifying a stronger global skill, but also from adapting
frame selection to different query categories.
The oracle category-to-skill mapping reaches 60.4\%,
indicating additional headroom for category-guided routing
within the current skill toolbox.
AutoSkill also matches the 58.6\% accuracy of the post-hoc
best fixed skill, which requires evaluating every toolbox
skill using target annotations.
In contrast, AutoSkill learns the category-to-skill mapping
without target videos or answers.
The Discovery Cycle Router reaches 57.4\%, while
majority voting also attains 58.6\% but requires five
video-MLLM passes per query. AutoSkill achieves
comparable performance with a single pass.

\noindent\textbf{Effect of routing set size.}
To measure how many samples are needed for routing table construction, we vary the routing-set size while keeping the discovered Top-5 toolbox and the inference configuration fixed. As shown in Table~\ref{tab:routing_size}, AutoSkill is already effective with a small set. A set of 300 samples gives a $+2.0\%$ average gain over the baseline. Increasing the size further improves performance gradually. The gain reaches $+2.2\%$ around 400--600 samples and $+2.3\%$ at 800--1{,}200 samples. With 1{,}500 samples the average reaches $+2.4\%$, and additional samples bring only marginal returns. We also randomly resample and rewrite the 1{,}500-example routing set five times and rebuild the table independently. The variation across seeds is negligible, which shows that the learned mapping is stable and not tied to one rewrite. We therefore use $N{=}1{,}500$ in the main experiments as a trade-off between routing reliability and construction cost.

\noindent\textbf{Why voting is not an effective router.}
Majority voting over the Top-5 skills reaches 58.6\%, matching the best
fixed skill but remaining below AutoSkill, while requiring five video-MLLM
passes per query. For many instances, only one or two skills recover the
critical evidence, making the correct answer a minority vote that is
overridden by mismatched skills. AutoSkill instead routes each query to its
specialised skill, preserving specialisation with only one pass.

\begin{table}[!t]
\caption{Routing-set size ablation on Qwen2.5-VL-7B (mean over 5 random rewrites from selected target-style samples). LongVB denotes LongVideoBench.}
\label{tab:routing_size}
\centering
\small
\vspace{-5pt}
\setlength{\tabcolsep}{1pt}
\begin{tabular}{c|cc|c|c|c|c}
\toprule
\multirow{2}{*}{Samples}
& \multicolumn{2}{c|}{MLVU}
& LongVB
& VideoMME
& LVBench
& Avg/$\Delta$ \\
\cmidrule(lr){2-3}
& dev & test & val & & & \\
\midrule
baseline
& 66.0 & 47.7 & 61.2 & 64.6 & 42.5 & 56.4 / -- \\
300
& 68.1 & 50.8 & 61.7 & 65.2 & 46.2 & 58.4 / +2.0 \\

400
& 68.5 & 51.2 & 61.7 & 65.1 & 46.3 & 58.6 / +2.2 \\

500
& 68.6 & 51.1 & 61.9 & 65.2 & 46.4 & 58.6 / +2.2 \\

600
& 68.6 & 51.1 & 61.8 & 65.2 & 46.6 & 58.6 / +2.2 \\

800
& 68.7 & 51.2 & 61.8 & 65.2 & 46.5 & 58.7 / +2.3 \\

1000
& 68.8 & 51.1 & 61.8 & 65.2 & 46.6 & 58.7 / +2.3 \\

1200
& 68.9 & 51.2 & 61.8 & 65.1 & 46.7 & 58.7 / +2.3 \\

1500
& \textbf{69.0} & \textbf{51.1}
& \textbf{61.9}
& \textbf{65.2}
& \textbf{46.8}
& \textbf{58.8 / +2.4} \\
\bottomrule
\end{tabular}
\vspace{-5pt}
\end{table}

\vspace{-5pt}
\section{Conclusion}
We presented AutoSkill, which replaces a universal frame selector with an
automatically discovered skill toolbox and taxonomy-guided routing.
An LLM agent develops complementary skills on a capability-focused development
set and adapts them to target domains without target answers or weight updates.
Experiments show that AutoSkill exploits skill complementarity to outperform
fixed-skill baselines, providing an effective training-free approach to
LVQA.

\bibliography{ref}

\end{document}